\documentclass[a4paper, 10pt, conference]{ieeeconf}      

\IEEEoverridecommandlockouts                              

\usepackage{amsmath} 
\usepackage{amssymb}  
\usepackage{graphicx}
\graphicspath{{images/}}
\usepackage[adjust]{cite}

\usepackage{pgfplots}
\pgfplotsset{compat=1.8}
\usepackage{tikzscale}

\usepackage{xcolor}
\usepackage{xspace}
\usepackage{csquotes}

\usepackage{caption}
\usepackage{subcaption}

\usepackage{algorithm}
\usepackage[noend]{algpseudocode}
\algrenewcommand\algorithmicrequire{\textbf{Given:}}
\algrenewcommand\algorithmicensure{\textbf{Wanted:}}

\usepackage{mathabx}

\usepackage{siunitx}
\DeclareSIUnit{\fps}{fps}

\usepackage{paralist}

\usepackage[hidelinks]{hyperref}
\usepackage{placeins}

\title{\LARGE \bf
Leveraging swarm capabilities to assist other systems}

\author{Miquel Kegeleirs$^{1}$, David Garzón Ramos$^{1}$, Guillermo Legarda Herranz$^{1}$, Ilyes Gharbi$^{1}$, \\Jeanne Szpirer$^{1}$, Ken Hasselmann$^{2}$, Lorenzo Garattoni$^{3}$, Gianpiero Francesca$^{3}$, and Mauro Birattari$^{1}$
\thanks{$^{1}$IRIDIA, Université libre de Bruxelles. 
$^{2}$Royal Military Academy of Belgium. 
$^{3}$Toyota Motor Europe.
Correspondence to mauro.birattari@ulb.be.}%
\thanks{The project has received funding from Belgium’s Wallonia-Brussels Federation through an ARC Advanced Project 2020 (Guaranteed by Optimization). MK and MB acknowledge support from the Belgian Fonds de la Recherche Scientifique-FNRS. DGR acknowledges support from the Colombian Ministry of Science, Technology and Innovation – Minciencias.} 
\thanks{MK, DGR, GLH, IG and JS contributed equally to this work and should be recognized as co-first authors. The idea was conceived by all authors. The experiments were designed by MK, DGR, GLH, IG and JS, and performed by MK with the assistance of DGR, GLH, IG and JS. The paper was drafted by MK, GLH and KH and edited by MB; 
The research was directed by MB.}%
}

\begin{document}

\maketitle
\thispagestyle{empty}
\pagestyle{empty}


Most studies in swarm robotics treat the swarm as an isolated system of interest.
Investigations into heterogeneous swarms~\cite{DorFloGam-etal2013IEEERAM,DucDiPinGam2011SI} examine various types of robots working in conjunction, yet these too are considered collectively as a single entity. 
While there are instances where swarms are supported by external systems~\cite{ZhuOguHei2024arxiv}, it remains rare for swarms to act in service of another system.

We argue that the prevailing view of swarms as self-sufficient, independent systems limits the scope of potential applications for swarm robotics. 
Specifically, certain scenarios, such as search and rescue operations, might not derive substantial advantage from deploying a robot swarm as an autonomous solution. However, the assistance provided by a swarm could prove invaluable to human rescuers.

Robot swarms are uniquely poised for effective information acquisition. Their distributed nature allows them to rapidly collect environmental data and continuously update this information through peer-to-peer sharing. We refer to this collective data-gathering capability as ``swarm perception.''
In swarm robotics, extensive research has been conducted on collective behaviors~\cite{TriCam2015hci,BraFerBirDor2013SI} and collective decision-making~\cite{StrCasDor2018aamas,ValFerHam-etal2016AAMAS}, often highlighting the importance of swarm perception.
For example, the works of Valentini et al.~\cite{ValBraHamDor2016ants} and Zakir et al.~\cite{ZakDorRei2022ants}, use swarm perception as a tool in a study evaluating collective decision-making.

Swarm perception enables a swarm to dynamically monitor its surroundings, detecting and reporting changes, even in environments whose structure is initially unknown.
While the information gathered is typically utilized internally to refine the swarm's collective behavior, it can also be seen as a vast, evolving shared database filled with mission-specific environmental data.
By granting external systems access to this database---for instance, through communication between swarm robots and an external robot---the swarm can provide critical information that aids the external system's task. 
However, there is a notable scarcity of research focused on leveraging swarm perception to benefit external systems. 

The potential beneficiaries of swarm support extend beyond artificial systems. 
Indeed, the notion of a swarm aiding another system has already been explored within the realm of human-swarm interaction~\cite{KolWalCha-etal2016IEEETHMS,HusAbb2018smc}: the human operator can leverage the information gathered by the swarm to improve operational efficiency or safety. 
For example, a rescuer could utilize the swarm's data to pinpoint victims' locations, or a speleologist might depend on maps created by the swarm for planning explorations. 
The concept of a robot swarm supporting firefighters has been previously addressed by Naghsh et al.~\cite{NagGanTan-etal2008roman}. 
However, research on human-swarm interaction remains limited to date.

We anticipate that a heterogeneous architecture comprising a robot swarm would be especially valuable for tasks such as identifying and tracking targets (specifically people), surveillance, and scouting activities.
Tracking individuals across multiple cameras, known as multi-target, multi-camera tracking (MTMCT), poses a significant challenge~\cite{AmoSebIzh-etal2023NEUCOM}. 
Person re-identification (Re-ID) techniques are developed to address these challenges,  particularly for tracking pedestrians across multiple non-overlapping camera views~\cite{TanAndAndSch2017cvpr,RisTom2018cvpr,GaiKar2021JRTIP}.
Person Re-ID often relies on low-resolution CCTV cameras; this necessitates the extraction of whole-body features, a process that is  affected by variables such as occlusions, lighting conditions, body poses, and clothing alterations~\cite{YeSheLin-etal2022}.
The small size and mobility of individual robots in a swarm afford them the ability to access diverse viewpoints within an environment, enabling close-up observations from multiple angles. 
This versatility facilitates the implementation of more effective identification systems, such as facial recognition based on combined observations to enhance detection accuracy.
Furthermore, in scenarios where environments are unknown and strategies dependent on fixed sensor placements or path planning are impractical~\cite{RobLac2016AR}, identifying and tracking people become exceedingly complex. 

In contrast, robot swarms can continuously share information about identified individuals, enabling effective tracking as individuals move through the environment. 
This shared intelligence allows for real-time updates on the locations of identified individuals, enhancing the swarm's capability to support the navigation of other robots within the system. 
For instance, a hospital robot tasked with delivering medications or supplies could utilize the swarm's data to pinpoint and navigate directly to the requesting doctor's location, showcasing the practical utility of swarm-assisted operations in dynamic environments.
In surveillance or monitoring, it is typical to deploy multiple CCTV cameras that relay information back to a central surveillance unit. 
Robot swarms could complement traditional CCTV systems and offer a significant enhancement to such traditional setups by swiftly communicating any incidents to the security or monitoring team, including those that occur within the cameras' blind spots.
Crucially, swarm robots possess the capability to not just detect but actively manage events. 
Unlike passive camera systems, a robot swarm can initiate immediate, localized responses to mitigate incidents as they await the arrival of specialized personnel, or they might even resolve the situation independently. 

As early as 2006, Schmickl et al.~\cite{SchMosCra2006iwsr} envisioned a scenario in which a robot swarm could gather environmental data to create a map.
This foresight has since evolved into the development of swarm-based simultaneous localization and mapping (swarm SLAM), a sophisticated form of swarm perception.
Despite its potential~\cite{KegGriBir2021FRAI}, it was only recently that Lajoie and Beltrame~\cite{LajBel2023arxiv} introduced the first swarm SLAM framework by integrating existing SLAM technologies---specifically RTABMap~\cite{LabMic2019JFR} and GTsam~\cite{gtsam})---with an innovative, decentralized technique for prioritizing inter-robot loop closures.
However, the current implementation of this method primarily generates pose graphs and demands structured navigation characterized by long, straight movements---conditions that are not typically associated with swarm behavior, which often relies on random walks. 
Generally, robot swarms excel at creating abstract or coarse maps swiftly rather than detailed occupancy grids, a trait that somewhat restricts their utility within the swarm itself.
Nevertheless, this mapping capability is exceptionally well-suited to scouting tasks, where the swarm's objective is to relay information to another system. 
In such applications, a robot swarm can rapidly produce a basic map of a vast, unexplored area. 
This preliminary mapping provides valuable navigational assistance to other robots or offers a comprehensive overview for human operators, showcasing the unique strengths of swarm SLAM in exploratory and reconnaissance missions.

While robot swarms hold promise as support mechanisms for other systems, several challenges hinder their widespread adoption.
First, the exploration into enabling technologies for robot swarms, such as SLAM and computer vision, is a relatively recent development. 
Although some initial successes have been recorded, the reliability of current solutions does not yet meet the standards required for practical applications.

Moreover, the field of swarm robotics has been limited by a scarcity of experiments conducted in real-world environments. 
Much of the research to date has been confined to simulations or carried out in simplistic settings using robots like the E-Puck~\cite{MonBonRae-etal2009arsc} and the Kilobot~\cite{RubAhlNag2012icra}, which are designed specifically for research.
While these platforms are cost-effective and suitably compact for large-scale experiments, their capabilities fall short of what is needed for realistic environmental testing.
Recently developed, more advanced platforms such as the S-Drone ~\cite{OguAllZhu-etal2022techrep}, the Mercator~\cite{KegTodGar-etal2022techrep} or the Summit XL~\cite{ArrSec2023acta}, offer enhanced capabilities but have yet to see widespread adoption.
There is also uncertainty regarding whether findings from existing research platforms will be applicable to the advanced robots necessary for real-world tasks. 
Despite the success of automatic design methods in simulations, they often struggle with the reality gap~\cite{JakHusHar1995ecal,HasLigRudBir2021NATUCOM}, making it challenging to anticipate how these methods will perform with more sophisticated robots. 
However, approaches like AutoMoDe~\cite{BirLigBoz-etal2019FRAI,BirLigFra2021admlsa} have shown some promise in bridging the reality gap~\cite{FraBraBru-etal2014SI,LigBir2020SI}, and the broader transfer gap~\cite{KegGarHas-etal2024ral}.

Another critical consideration is the secure storage and sharing of information collected by the swarm~\cite{HunHau2020NATUMINT}.
The inherent decentralized and distributed nature of robot swarms provides a measure of data security, as information is not centralized. 
Nonetheless, practical applications may necessitate reliance on more centralized and potentially vulnerable infrastructures, especially when swarms need to share data with other systems. 
Additionally, robot swarms could be susceptible to various attack vectors, including the threat of byzantine robots~\cite{StrPacDor2023SCIROB}.
Blockchain technology has been identified as a potential solution to offer transparency, security, and trust within robot swarms~\cite{DorPacRei-etal2024NATUREE}.

\begin{figure}
    \centering
    \includegraphics[width=.98\columnwidth]{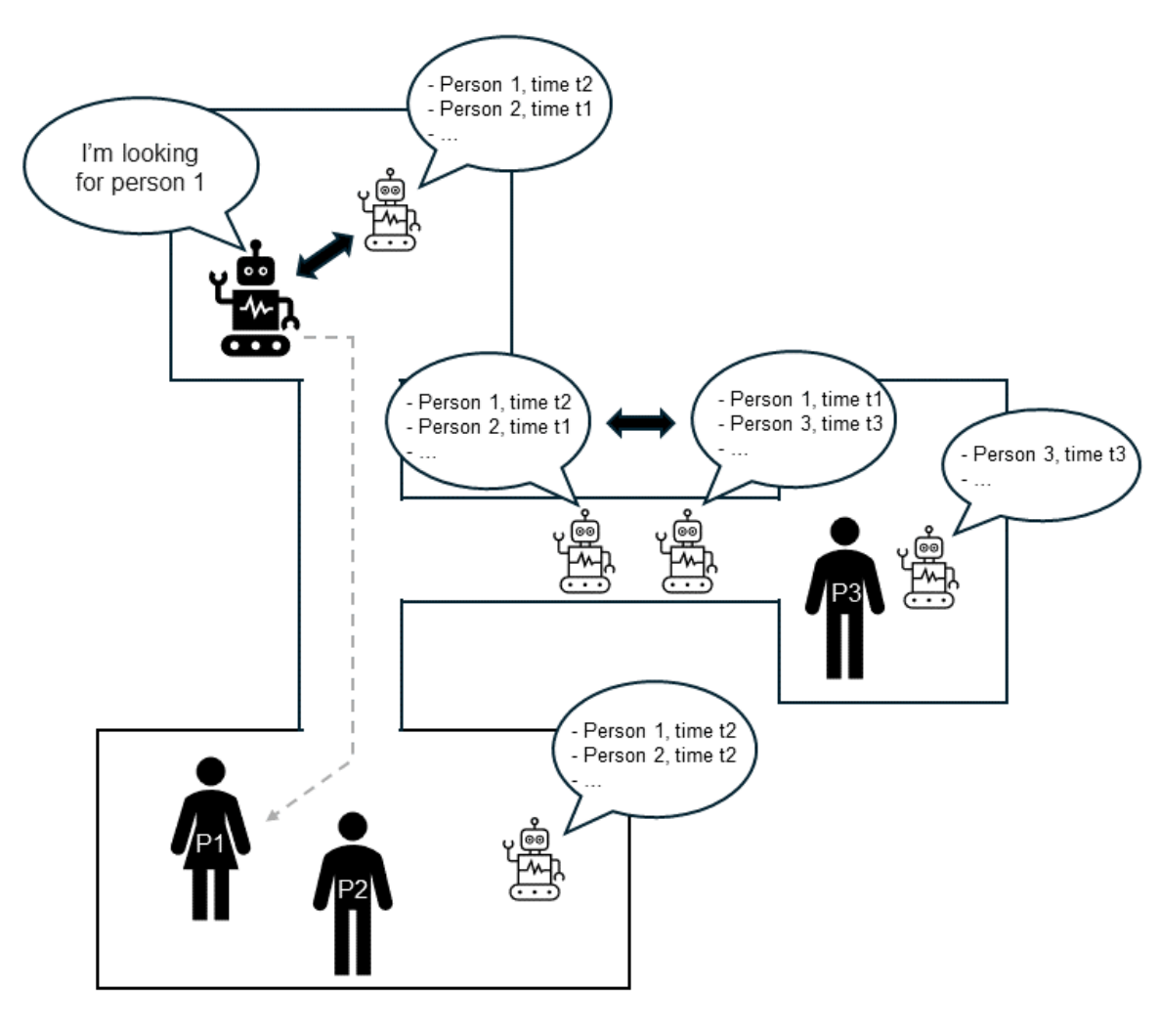}
    \caption{A swarm of robots, depicted in white, navigates through an office setting, identifies individuals, and logs both their locations and the times of detection. This information is shared among the swarm to maintain updated records and is also communicated to another robot, shown in black, that requests this information.}
    \label{fig:mission}
\end{figure}

To overcome these obstacles, collaboration with industry could play a crucial role by aligning research with practical needs. 
With the support of Toyota Motor Europe, we realized a demonstration of the ideas presented in this article.
The demonstration\footnote{The video, also submitted to this workshop, is available at \href{https://www.youtube.com/watch?v=8vkeuyJllY8}{https://www.youtube.com/watch?v=8vkeuyJllY8}.}, showcases a swarm of Mercators~\cite{KegTodGar-etal2022techrep} aiding a Toyota Human Support Robot (HSR)  in delivering an item to a person in an unknown location. 
The concept is illustrated in Figure~\ref{fig:mission}.
Our vision is that further experiments in this direction would open the field of swarm robotics to new perspectives and eventually enable  concrete applications.
\FloatBarrier

\bibliographystyle{IEEEtran}
\bibliography{demiurge-bib/definitions,demiurge-bib/author,demiurge-bib/address,demiurge-bib/proceedings-short,demiurge-bib/journal-short,demiurge-bib/publisher,demiurge-bib/series-short,demiurge-bib/institution,demiurge-bib/bibliography,additions}

\end{document}